\pdfoutput=1

\documentclass[11pt]{article}
\usepackage[final]{acl}

\usepackage{colortbl}
\usepackage{amsmath}
\usepackage{arydshln}
\usepackage{amsmath}
\usepackage{times}
\usepackage{latexsym}
\usepackage{booktabs}
\usepackage{multirow}
\usepackage{xcolor}
\usepackage{colortbl}
\usepackage{graphicx}
\usepackage[T1]{fontenc}
\usepackage{algorithm}
\usepackage{algpseudocode}
\usepackage[utf8]{inputenc}
\usepackage{amsmath, amssymb}
\usepackage{microtype}
\usepackage{enumitem}
\usepackage{makecell}
\usepackage{inconsolata}

\usepackage{graphicx}

\newtheorem{definition}{Definition}

\newcommand{\Input}{\item[\textbf{Input:}]}
\newcommand{\Output}{\item[\textbf{Output:}]}

\title{Image Corruption-Inspired Membership Inference Attacks against \\ Large Vision-Language Models}

\author{Zongyu Wu, Minhua Lin, Zhiwei Zhang, Fali Wang, Xianren Zhang,\\
\textbf{Xiang Zhang, Suhang Wang} \\
The Pennsylvania State University\\
\texttt{\{zongyuwu,mfl5681,zbz5349,fqw5095,xzz5508,xzz89,szw494\}@psu.edu}
}

\begin{document}
\maketitle
\begin{abstract}
Large vision-language models (LVLMs) have demonstrated outstanding performance in many downstream tasks. However, LVLMs are trained on large-scale datasets, which can pose privacy risks if training images contain sensitive information. Therefore, it is important to detect whether an image is used to train the LVLM. Recent studies have investigated membership inference attacks (MIAs) against LVLMs, including detecting image-text pairs and single-modality content. In this work, we focus on detecting whether a target image is used to train the target LVLM. We design simple yet effective Image Corruption-Inspired Membership Inference Attacks (ICIMIA) against LVLMs, which are inspired by LVLM's different sensitivity to image corruption for member and non-member images. We first perform an MIA method under the white-box setting, where we can obtain the embeddings of the image through the vision part of the target LVLM. The attacks are based on the embedding similarity between the image and its corrupted version. We further explore a more practical scenario where we have no knowledge about target LVLMs and we can only query the target LVLMs with an image and a textual instruction. We then conduct the attack by utilizing the output text embeddings' similarity. Experiments on existing datasets validate the effectiveness of our proposed methods under those two different settings.

\end{abstract}

\section{Introduction}

Large Vision-Language Models (LVLMs)~\citep{liu2024llava1.5,liu2024llavanext,bai2023qwenvl}, which can generate text outputs based on visual and/or textual input, are attracting increasing attention. Many LVLMs have been developed, which have shown great performance on various tasks~\citep{dong2024benchmarking,xu2024gnnllm,wu2025lanp,yue2023mmmu,li2024seedbenchplus,bucciarelli2024personalizing} such as biomedical question answering~\citep{li2023medllava}. Despite their great success, LVLMs have also raised critical concerns about privacy and copyright issues. LVLMs are usually trained on large-scale datasets~\citep{changpinyo2021conceptual,kakaobrain2022coyo-700m}. The training data might contain sensitive information, such as unauthorized medical data and copyrighted content. Previous works have shown that neural networks, especially those trained on large-scale datasets, can memorize training data~\citep{song2017machine,carlini2019secret,carlini2023quantifying}. Thus, the memorization phenomena might also happen in LVLMs and inadvertently cause training data leakage~\citep{LiWCTAC24}, which could cause substantial loss to data owners. Hence, knowing whether one's data is used to train an LVLM is important for privacy and copyright protection.

Membership inference attacks (MIAs), which aim to determine whether a given sample is used to train a model~\citep{hu2022miasurvey,shokri2017miaml}, are critical for ensuring data safety and investigating data contamination~\citep{oren2024proving,duan2024membership}. Generally, models tend to overfit to the training data, resulting in higher prediction confidence for member data (data used for training) than non-member data. Many traditional MIA methods~\citep{shokri2017miaml} adopt such nuance difference in model prediction to differentiate members and non-members. Recently, some works~\citep{ko2023practical,LiWCTAC24,hu2025membership} have explored the MIA on vision-language models, from CLIP~\citep{radford2021clip} to LVLMs~\citep{zhang2023llamaadapter}. \citet{ko2023practical} focus on determining whether an image-text pair exists in the training data of CLIP. However, detecting whether an image in the training data is more practical than detecting the entire image-text pair~\citep{LiWCTAC24} as the image owner might only have the image without text while the text description might be labeled by the model trainer.

Therefore, we study the problem of single-image MIA against LVLMs. The work in this direction is rather limited~\cite{LiWCTAC24}. \citet{LiWCTAC24} also conduct MIA on a single modality (Image or textual description) against LVLMs by using the output logits of LVLM. Instead of using output logits, we propose a new perspective, i.e., adopting image embedding from the visual part of LVLMs. Our intuition is: as the model has seen member images, it should be able to give robust image embedding of a member image even if some details of the image are missing. In other words, \textit{the image embedding of a member is more robust to image corruption than that of non-member images}, which is verified by our preliminary experiment in Section~\ref{sec:pre} (see Figure.~\ref{fig:sim_histogram}). Based on this observation, we propose a novel MIA algorithm under the white-box setting, where we can obtain the image embedding from LVLMs. Given an image, we corrupt it and use the image embedding similarity between the raw image and the corrupted version to decide if it is a member. A higher similarity means the image is more likely to be a member.

However, for many closed-source LVLMs, we cannot obtain image embeddings. To address this issue, we extend the similarity to the output text level. Our assumption is: robust image embedding of member image will result in robust text generation under perturbation, which is also verified in Figure~\ref{fig:sim_histogram_text}. Based on this observation, we extend our framework to black-box setting, where we can only obtain output texts from LVLMs. Given a target image, we corrupt the image and compare the generated text similarity between the raw image and its corrupted version. A larger text similarity means the image is more likely to be a member.

Our \textbf{main contributions} are: (\textbf{i}) In this work, we investigate two membership inference attack scenarios targeting LVLMs. For each setting, we propose one simple yet strong attack method that leverages the model’s robustness to image corruption on its training images; (\textbf{ii})  Extensive experiments on existing datasets show the effectiveness of the proposed membership inference method.

\section{Related Work}
\subsection{Large Vision-Language Models}
Large Vision-Language Models~\citep{liu2024llavanext,liu2024llava1.5,chen2024internvl,tong2024cambrian,zhu2024minigpt,chen2023minigptv2,li2024monkey,chen2024internvl15}, also known as multimodal large language models~\citep{fu2024mmevalsurvey} are developing rapidly due to the success of language models across various aspects~\citep{wang2024comprehensive,zhao2023survey,zhang2025divideandrefine,vicuna2023,touvron2023llama,touvron2023llama2,2023internlm,penedo2023refinedweb,lin2024decoding}. These multimodal models, like LLaVA series~\citep{liu2023llava,liu2024llava1.5}, can generate textual outputs given textual questions and images.

\subsection{Membership Inference Attack}
Membership Inference Attack (MIA)~\citep{shokri2017miaml,salem2018ml,sablayrolles2019white,li2021miadinclassify,hu2022miasurvey,nasr2019comprehensive,leino2020stolen,rezaei2021difficultymia} tries to determine whether a given data sample was used to train a machine learning model~\citep{shokri2017miaml}.

One main category of MIA methods is metric-based~\citep{sablayrolles2019white,choo2021labelmia}, which use some well-designed metrics~\citep{hu2022miasurvey}, such as the metrics based on loss, to determine the membership status of a given sample. Another main category consists of methods based on shadow training~\citep{shokri2017miaml}, which trains some shadow models to simulate the target model and then conducts MIAs based on these shadow models. Many later works~\citep{mireshghallah2022empirical,mattern2023mianeighbourhood,ren2024self,Min-K}, such as Min-K\%~\citep{Min-K} and Min-K\%++~\citep{zhang2025mink++}, have started to explore the MIA on Large Language Models (LLMs).

With the development of multimodal learning, there are also some works exploring MIA on multimodal models~\citep{hu2022m,ko2023practical,hu2025membership,LiWCTAC24}. We focus on vision-text here. EncoderMI~\citep{liu2021encodermi} studies MIAs on image encoder models such as CLIP vision encoder~\citep{radford2021clip}. It calculates the similarity scores between the augmented images' embeddings and the scores are then used to train a classifier to infer the member status of an image. Our work is inspired by EncoderMI. \citet{ko2023practical} work on detecting whether an image-text pair is in the training data of CLIP models. \citet{LiWCTAC24} investigate the single-modality MIA in LVLMs, which is a more practical scenario. They calculate the R{\'e}nyi Entropy~\citep{renyi1961measures} based on different slices of logits to infer membership.~\citet{hu2025membership} found that member data and non-member data have different sensitivity to temperature. They then perform four attack methods under four different settings to detect whether images are used in LVLM's training stage.

Our work is inherently different from existing work: We propose a new perspective from image embedding robustness and output text robustness of the member image under image corruption, which works for both white-box and black-box settings.

\section{Preliminaries}
\label{sec:pre}
In this section, we give the necessary background information and formulate the problems for MIAs against LVLMs.

\begin{figure}[t!] 
\centering 
\includegraphics[width=1\linewidth]{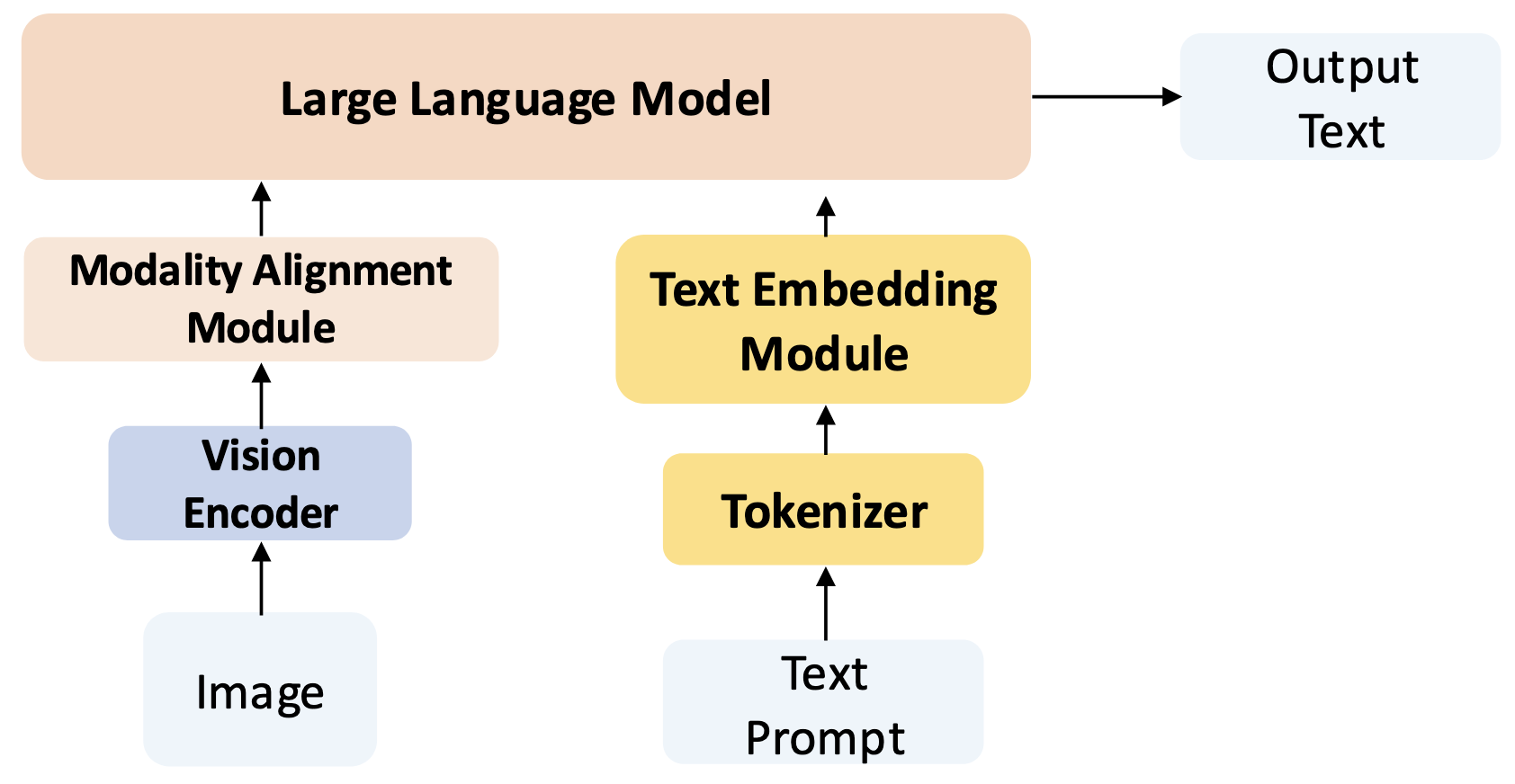}
\caption{An illustration of the architecture of large vision-language models~\citep{liu2024llava1.5}.} 
\label{fig:lvlm}
\end{figure}

\subsection{Large Vision-Language Models}
As shown in Figure~\ref{fig:lvlm}, an LVLM $\mathbf{M}_\theta$ usually consists of three parts~\citep{liu2024llava1.5}: A vision encoder $f_{Vision}$, an LLM $f_{LLM}$, and a modality connection module $f_{Align}$. The vision encoder $f_{Vision}$, e.g., CLIP~\cite{radford2021clip} vision encoder used in LLaVA-1.5, takes an image $\mathbf{x}$ as input and outputs the embeddings of $N$ image patches (excluding the CLS token) $\{\mathbf{z}_1,...,\mathbf{z}_N\}$, where $\mathbf{z}_i \in \mathbf{R}^{d_v}$ is the $i$-th patch embedding with $d_v$ being the embedding dimension. The patch embeddings are then transformed into an embedding sequence $\mathbf{E}^v = \{\mathbf{e}_1^v,...,\mathbf{e}_N^v\}$ which is in the embedding space of $f_{LLM}$ using $f_{Align}$. Each element in $\mathbf{E}^v$ can be represented as:
\begin{equation}
\label{eq:image_embedding}
    \mathbf{e}_i^v = f_{Align}(\mathbf{z}_i),
\end{equation}
The text prompt $T_{in}$, composed of the instruction and the question, is tokenized and then encoded into an embedding sequence of tokens $\mathbf{E}^t = \{\mathbf{e}_1^t,...,\mathbf{e}_K^t\}$, where $K$ is the number of tokens. Finally, the text embedding and the image embeddings are used as input to the LLM $f_{LLM}$ to get the text output $T_{out}$.

\begin{figure}[t!] 
\centering 
\includegraphics[width=1\linewidth]{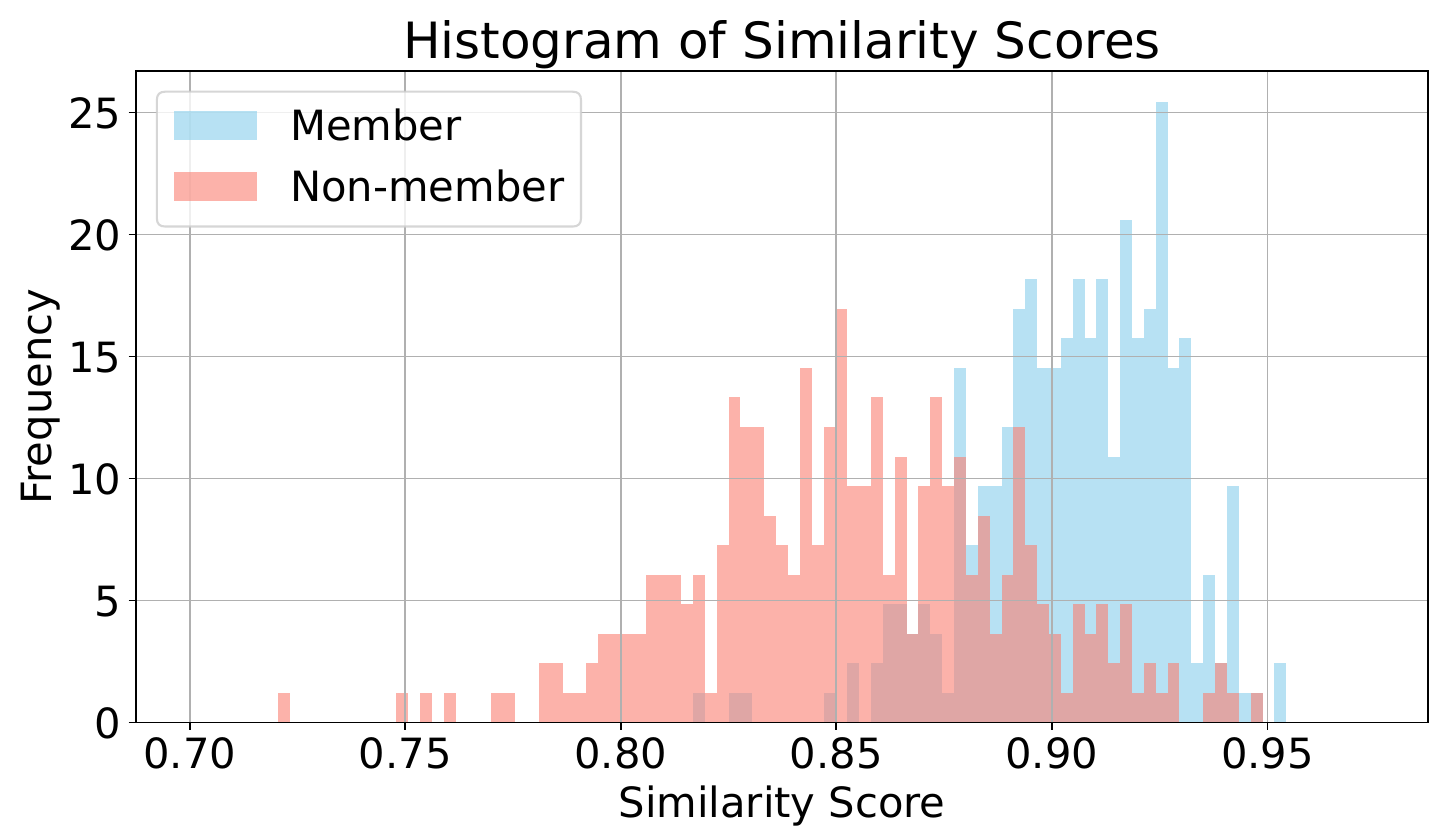}
\caption{A histogram of similarity scores of corrupted images' embeddings and original images' embeddings for member and non-member data.} 
\label{fig:sim_histogram}
\end{figure}

\begin{figure}[t!] 
\centering 
\includegraphics[width=1\linewidth]{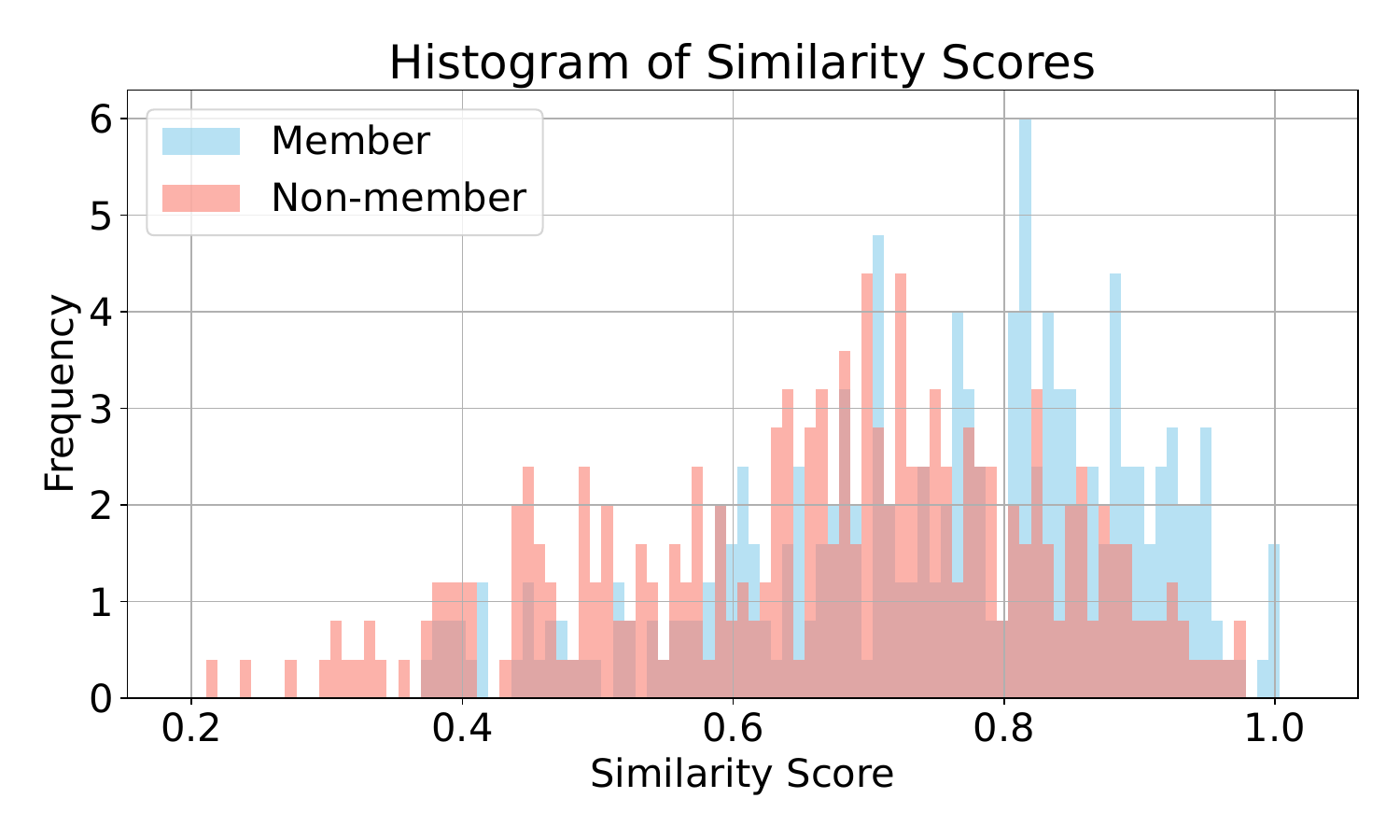}
\caption{A histogram of similarity scores of textual output embeddings between corrupted images and the original images for member and non-member data.} 
\label{fig:sim_histogram_text}
\end{figure}

\subsection{Threat Model}
\noindent\textbf{Attacker's Goal}. 
Given a trained LVLM $\mathbf{M}_\theta$, the attacker's goal is to determine if a specific image was in the training set of $\mathbf{M}_\theta$. 
The attackers only have the target image and do not require its corresponding ground-truth text description. The problem is formally defined as: 
\begin{definition}[Image Only MIA]
    Given a trained LVLM $\mathbf{M}_\theta$ and a target image $\mathbf{x_t}$, the attacker aims to determine whether $\mathbf{x_t}$ was part of the training data for $\mathbf{M}_\theta$, i.e., $\text{Attack}(\mathbf{M}_\theta, \mathbf{x_t}, T_{in}) \rightarrow \{0,1\} $, 
where $T_{in}$ is the input textual instruction. $\mathbf{x_t}$ is considered to be in the training set (i.e, member image) if the output is 1. Otherwise, $\mathbf{x}_t$ is a non-member.
\end{definition}

\noindent\textbf{Attacker's Knowledge}.
In our work, we consider two practical settings, i.e., white-box and black-box: (i) \textbf{White-box Setting.} In this setting, the attacker can get the embedding of the image, i.e., the embedding through the model's vision encoder and modality alignment module. This is practical for open-source LVLMs. $T_{in}$ is not required in this setting.
(ii) \textbf{Black-box Setting.} In this setting, the attacker can only query the  LVLM $\mathbf{M}_\theta$ with image input $\mathbf{x}$ and text input $T_{in}$ to get the text output $T_{out}$. The attacker has no knowledge of the target model. This setting is more realistic for commercial LVLMs.

\subsection{Intuition for Performing MIA}
\label{subsec:simdiff}
As we are only given the image, how to fully utilize the image to determine if the image remains a challenging question. 
Inspired by the findings of EncoderMI~\citep{liu2021encodermi} that a CLIP vision encoder tends to overfit to its training data and member images show higher similarity for two augmented versions, we explore whether LVLMs have similar features under image corruption methods. We assume that \textbf{the image embedding of a member image from the vision encoder and modality alignment module should be more robust to image corruption than that of non-member images}. The intuition is, if the model has seen the image and memorized the image, then even if the image is corrupted, e.g., some details missing, the model can still recall the details and result in embedding similar to the original one. To verify our assumption, for each image, we first apply \textit{Gaussian Blur} as the corruption method to obtain the corrupted image, where Gaussian Blur is a smoothing technique that reduces image detail and noise by averaging pixels with a weighted Gaussian kernel. We compute the similarity score between the embedding of the original image and that of its corrupted version. The chosen model is LLaVA-1.5-7B~\citep{liu2024llava1.5}, and the dataset is the VL-MIA/Flickr dataset constructed by~\citet{LiWCTAC24}. The similarity scores of member images and non-member images are shown in Figure~\ref{fig:sim_histogram}. We can observe that member images normally show a higher similarity score than non-member images, which aligns well with our assumption. The scores show a clear difference between the two groups, which suggests that the similarity score between the original image embedding and its corrupted version's embedding can be used to perform an image membership inference attack. Thus, we can utilize the robustness of member image embedding to determine its membership.

However, in the black-box setting, we are unable to obtain the image embedding. As image embedding of member image is robust, it might also result in robust text even if the image is corrupted. Thus, we assume that \textbf{given the same text prompt (Instruction), an image seen during the LVLM's training process will produce a text output that is more similar to the output generated from its corrupted version, compared to images not included in the training data}.  To verify our assumption, we use the same dataset, model, and corruption method as above. We compute the similarity score between the output text embeddings of the original and corrupted versions of each image, given the same prompt, separately. The results are shown in~Figure~\ref{fig:sim_histogram_text}. We observe that the discrepancy in similarity between member and non-member images still exists and can serve as a metric, although it is less apparent than that in Figure~\ref{fig:sim_histogram}.

\begin{figure*}[t] 
\centering 
\includegraphics[width=1\linewidth]{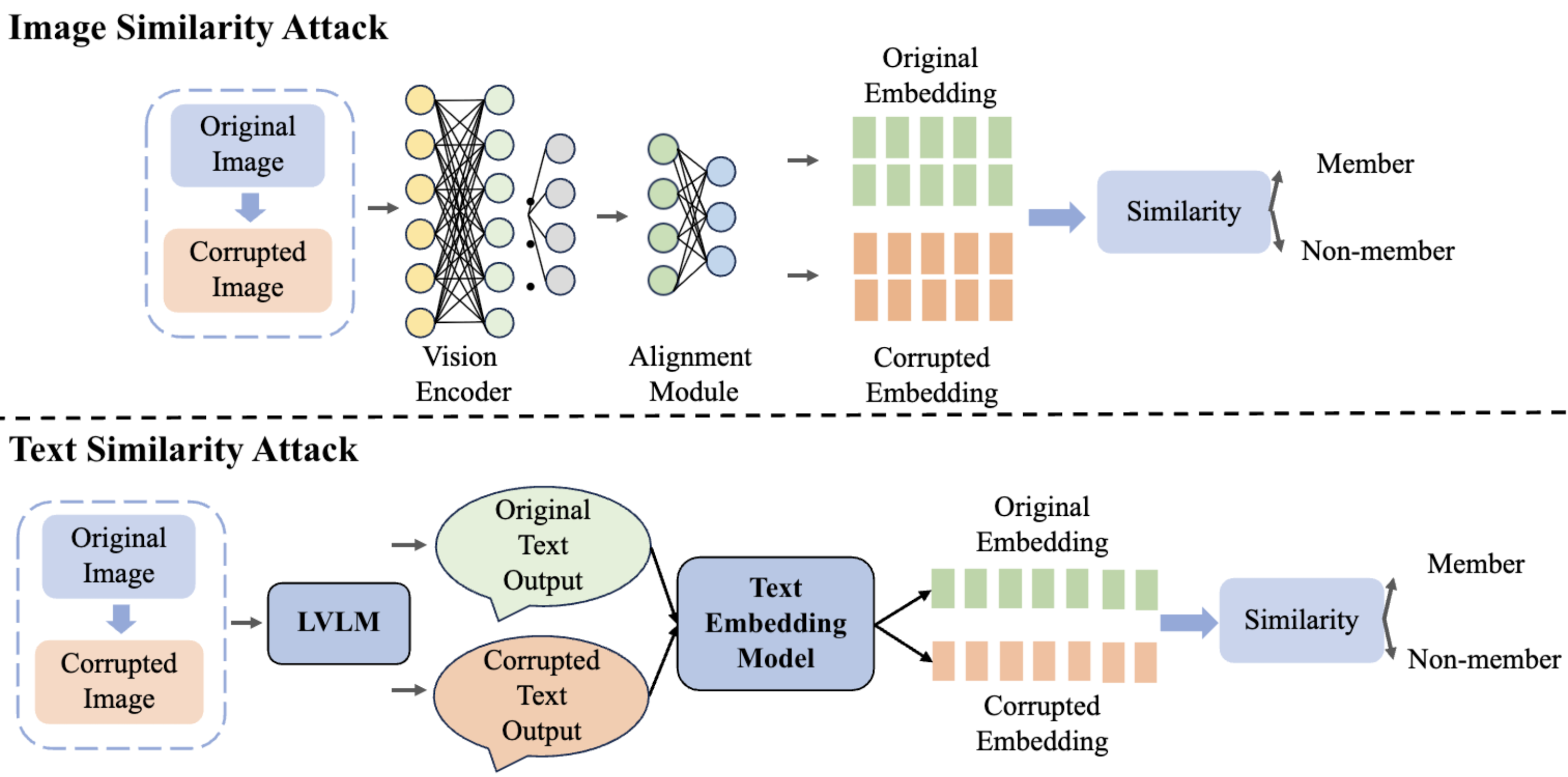}
\caption{An illustration of the attack pipeline under two different settings.} 
\label{fig:framework}
\end{figure*}

\section{Method}

Based on observations that LVLMs are more robust to corruptions on members than non-members, we propose a novel framework, ICIMIA (\textbf{I}mage \textbf{C}orruption-\textbf{I}nspired \textbf{M}embership \textbf{I}nference \textbf{A}ttacks against Large Vision-Language Models). As shown in Figure~\ref{fig:framework}, our method first produces a corrupted version $\mathbf{x}_t'$ for the target image $\mathbf{x}_t$. Then both the $\mathbf{x}_t$ and $\mathbf{x}_t'$ are utilized to get their image embeddings or corresponding output text embeddings. Finally, we calculate the image/text embedding similarity as a metric. The image is viewed as a member image if the similarity score is bigger than a certain threshold.  Next, we introduce the details.

\subsection{White-Box MIA via Image Similarity}

\begin{algorithm}[t]
\caption{Image Similarity-based MIA}
\label{algo:sim}
\begin{algorithmic}[1]
\Input Target image $\mathbf{x}_t$, threshold $\lambda$
\Output Membership Prediction Result $ \in \{0, 1\}$
\State Obtain the embedding $\mathbf{E}^v$ of $\mathbf{x}_t$
    \State Apply corruption on $\mathbf{x}_t$ to get image $\mathbf{x}_t'$
    
    \State Obtain the embedding $\mathbf{E}^{v'}$ of $\mathbf{x}_t'$
      \State Compute similarity score
    $ s_{img}$ via Equation~\ref{eq:simimage}
\If{$s_{img} < \lambda$}
    \State  $\mathbf{x}_t$ is a viewed as a non-member image
\Else
    \State $\mathbf{x}_t$ is a viewed as a member image
\EndIf
\end{algorithmic}
\end{algorithm}

We first discuss the white-box setting where we can obtain the embeddings of a given image through the vision encoder and modality alignment module. 

Based on our observation in Section~\ref{subsec:simdiff}, the embeddings of trained images should be more robust to image corruption methods. To be specific, compared to images not used during training (Non-member data), the embeddings of the corrupted image and its original counterpart are normally more similar for images that were used in the training process (Member data). Therefore, we can use the similarity score as a metric to determine whether an image is used to train the LVLM.

The pipeline of our proposed framework is shown in the upper part of Figure~\ref{fig:framework}. For a target image $\mathbf{x}_t$ that we wanna detect, we first apply some corruptions to get a corrupted image as $\mathbf{x}_t' = \text{Corruption}(\mathbf{x}_t)$. Various corruption techniques can be used, such as Gaussian blur (Using a Gaussian Kernel). Then, we get the image embeddings for both the original image and the corrupted one. Last, we measure how close each patch embedding pair is and calculate the mean value as
\begin{equation}
\label{eq:simimage}
    s_{img} = \frac{1}{N} \sum_{i=1}^{N} \text{Sim}(\mathbf{e}_i^v, \mathbf{e}_i^{v'})
\end{equation}
where $\mathbf{e}_i^v$ is the embedding of the $i$-patch of image $\mathbf{x}_t$ obtained by Equation~\ref{eq:image_embedding} and $\mathbf{e}_i^{v'}$ denotes the $i$-patch embedding of the corrupted image $\mathbf{x}_t'$. $N$ is the number of patches. $\text{Sim}(,)$ is a similarity function. We use cosine similarity here. The larger the $s_{img}$ is, the more likely the image is viewed as a member image. The target image $\mathbf{x}_t$ is predicted as a member image if $ s_{img}$ is bigger than a threshold $\lambda$. Algorithm~\ref{algo:sim} summarizes the image similarity-based attack.

\subsection{Black-box MIA via Text Similarity}
For many commercial LVLMs, we are not able to obtain image embeddings. Thus, we study a more practical black-box setting where we know nothing about the target model but can only query the model with the target image and prompt to obtain the response text. Though we cannot obtain the image embedding, as the image embeddings of member images are robust to random corruption, correspondingly, the generated response text will be robust to the corruption, which is also verified in~\ref{subsec:simdiff}. This motivates us to query the target model using the original image and its corrupted version. Then we calculate pair-wise output text similarities. 

Specifically, we feed the original target image $\mathbf{x}_t$ and a text prompt $T_{in}$ into the target LVLM and get an output $T_{out}$ as $T_{out} = f_{LVLM}(\mathbf{x}_t, T_{in})$. Similarly, we get an output $T_{out}'$ for the corrupted image $\mathbf{x}_t'$ with the same input text prompt $T_{in}$. We then employ a text embedding model to get the embeddings of $T_{out}$ and $T_{out}'$. With the text embedding, we calculate their similarity as
\begin{equation}
\label{eq:textsim}
   s_{text} = \text{Sim}(Emb(T_{out}),Emb(T_{out}'))
\end{equation}
where $Emb()$ denotes a text embedding model, such as OpenAI's text-embedding-3-small model\footnote{https://platform.openai.com/docs/models}. $\text{Sim}(,)$ is a similarity function as in Equation~\ref{eq:simimage} which is also a cosine similarity function. Similarly, $x_t$ is considered member data if \( s_{text} > \lambda \). This method is summarized in Algorithm~\ref{algo:blackbox}. This attack is similar to the Target-only Inference in~\citet{hu2025membership} where they also use text similarity scores. The difference is that they use the average similarity score between text outputs generated by querying the model multiple times.

\begin{algorithm}[t]
\caption{Text Embedding Similarity-based Attack}
\label{algo:blackbox}
\begin{algorithmic}[1]
\Input Target image $\mathbf{x}_t$, Prompt $T_{in}$, threshold $\lambda$
\Output Membership Prediction Result $ \in \{0, 1\}$
    \State Feed the target LVLM with the target image $\mathbf{x}_t$ and a text prompt $T_{in}$ and get output $T_{out}$
    \State Apply corruption to get corrupted image $\mathbf{x}_t'$ 
        \State Feed the target LVLM with the corrupted target image $\mathbf{x}_t'$ and the same text prompt $T_{in}$ and get output $T'_{out}$
      \State Get the text output embedding $Emb(T_{out})$ and $Emb(T_{out}')$ of the original image and corrupted image 
      \State Calculate the output similarity using Equation~\ref{eq:textsim}
\If{$  s_{text} < \lambda$}
    \State $\mathbf{x}_t$ is viewed as a non-member image
\Else
    \State $\mathbf{x}_t$ is viewed as a member image
\EndIf
\end{algorithmic}
\end{algorithm}

\section{Experiments}
In this section, we evaluate our proposed ICIMIA on representative LVLM MIA datasets to answer these research questions: (\textbf{i}) How well do our proposed ICIMIA perform in conducting membership inference attacks against large vision-language models? and (\textbf{ii}) What are the impacts of hyperparameters?

\subsection{Experimental Setup}
\textbf{Evaluation Metric.} Following previous work~\cite{LiWCTAC24}, we use Area Under the Curve \textbf{(AUC)} and True Positive Rate at 5\% False Positive Rate \textbf{(TPR at 5\% FPR)} as the evaluation metrics. For both metrics, higher values indicate better MIA performance. The descriptions of these metrics can be found in Appendix~\ref{app:metric}.

\noindent\textbf{Datasets}
We test our method on two benchmark datasets, VL-MIA/Flickr and VL-MIA/Flickr-2k~\citep{LiWCTAC24}. The details of datasets can be found in Appendix~\ref{app:datasets}. 

\noindent\textbf{Computational Resources.}
We conduct all experiments on machines equipped with NVIDIA RTX A6000 GPUs (48GB memory each).

\noindent\textbf{Selected Models.}
We evaluate our method on two models: LLaVA 1.5 7B, and LLaVA 1.5 13B~\citep{liu2024llava1.5}. The models are chosen as they are classical and the datasets are applicable to them.

\noindent\textbf{Corruption Methods.}
We choose the following corruption methods:
(i) \textit{Gaussian Blur}: This is a technique that makes image soft and blurry using a Gaussian Kernel; (ii) \textit{Motion Blur}: We apply a custom convolution kernel, mimicking the effect of motion-blur; and (iii) \textit{JPEG Compression}: This is a method to compress the image to get the corrupted version.  We apply OpenCV~\citep{opencv_library} to achieve Gaussian Blur and Motion Blur.

\noindent\textbf{Baselines.}
We adopt representative and state-of-the-art MIA methods against LVLMs, including: (i) \textbf{AugKL}~\citep{LiWCTAC24,liu2021encodermi}:~\citet{LiWCTAC24} extend the approach designed by \citet{liu2021encodermi} to LVLMs. Specifically, \citet{LiWCTAC24} quantify the difference between the original and augmented images (Such as Crop and Rotation) by computing the KL divergence between their distributions of logits; (ii) \textbf{Max\_Prob\_Gap}~\citep{LiWCTAC24}: It is the average value of the difference between the highest and second-highest token probabilities at each position; (iii) \textbf{Min-K\%}~\citep{Min-K}: Min-K is an MIA method designed for LLMs. It uses a ground-truth token and computes the lowest K\% of its predicted probabilities. \citet{LiWCTAC24} extends it to the LVLM domain; (iv) \textbf{Perplexity}: It is based on loss.~\citet{LiWCTAC24} analyze target perplexity to achive the attacks~\citep{carlini2021extracting}; (v) \textbf{MaxRényi-K\%}~\citep{LiWCTAC24}: It selects the top K\% tokens with the highest Rényi entropy. Then the value is the average value of these entropies; and
(vi) \textbf{Mod-Rényi}~\citep{LiWCTAC24}: This is an extended version of MaxRényi-K\% and is designed for target-based scenarios.

\begin{table*}[ht]
    \centering
    \resizebox{\textwidth}{!}{
    \begin{tabular}{lcccccccc}
        \toprule
        \multirow{2}{*}{\textbf{Method}} & \multicolumn{4}{c}{\textbf{LLaVA-1.5-7B}} & \multicolumn{4}{c}{\textbf{LLaVA-1.5-13B}} \\
        \cmidrule(lr){2-5} \cmidrule(lr){6-9}
        & img & inst & desp & inst+desp & img & inst & desp & inst+desp \\
        \midrule
        Perplexity
        & - & 0.378 & 0.665  &  0.558
         & - & 0.440 & 0.707 & 0.646\\
         
         Min\_20\% Prob & - & 0.374 &  0.672 &  0.370
         & -& 0.454 & 0.684 & 0.433 \\

        \text{ModR\'enyi}& - & 0.370 & 0.658  &  0.613
         & - & 0.442 & 0.703 & 0.678\\
         
        Max\_Prob\_Gap 
        & 0.579 & 0.605 & 0.644 & 0.645 
         & 0.565& 0.501 & 0.656 & 0.652  \\

        Aug\_KL 
        & 0.665 & 0.568 & 0.537  &  0.549
         & 0.636& 0.540 & 0.538 &  0.552\\
         
    \text{MaxR\'enyi} 
    &  0.702& 0.726 & 0.709  &  0.743
         & 0.647 & 0.682 & 0.728 &  0.738\\
        \midrule

    \rowcolor{gray!15} \textbf{Ours} (Image\_Sim, Gaussian Blur, Kernel Size 5) & \multicolumn{4}{c}{0.881}  & \multicolumn{4}{c}{0.878}  \\

        \rowcolor{gray!15} \textbf{Ours} (Image\_Sim, Motion Blur Kernel Size 5) & \multicolumn{4}{c}{0.860}  & \multicolumn{4}{c}{0.856}  \\

        \rowcolor{gray!15} \textbf{Ours} (Image\_Sim, JPEG Compression, Quality = 5) & \multicolumn{4}{c}{0.682}  & \multicolumn{4}{c}{0.681}  \\
   
        \bottomrule
    \end{tabular}
    }
    \caption{\textbf{AUC} of various baseline methods under \citet{LiWCTAC24}'s pipeline and our proposed approach on VL-MIA/Flickr. For all the baselines, the term "img" refers to the logits segment associated with the image embedding, while "inst" represents the instruction segment~\citep{LiWCTAC24}. "desp" corresponds to the generated description's logits segment, and "inst+desp" denotes the combination of the instruction and description segments~\citep{LiWCTAC24}.}
        \label{tab:vl-mia-auc}
\end{table*}

\begin{table*}[htbp]
    \centering
    \resizebox{\textwidth}{!}{
        \begin{tabular}{lcccccccc}
        \toprule
        \multirow{2}{*}{\textbf{Method}} & \multicolumn{4}{c}{\textbf{LLaVA-1.5-7B}} & \multicolumn{4}{c}{\textbf{LLaVA-1.5-13B}} \\
        \cmidrule(lr){2-5} \cmidrule(lr){6-9}
        & img & inst & desp & inst+desp & img & inst & desp & inst+desp \\
        \midrule

   Perplexity
        & - & 0.007 & 0.137  &  0.067
         & - & 0.047 & 0.227 & 0.127\\
         
         Min\_20\% Prob & - & 0.007 &  0.127 &  0.003
         & -& 0.067  &  0.163 & 0.053  \\

        \text{ModR\'enyi}& - & 0.003 & 0.113  &  0.113
         & - &  0.060 &  0.203  & 0.147 \\
         
        Max\_Prob\_Gap 
        & 0.050 & 0.083 & 0.163 & 0.163 
         & 0.050  & 0.107  & 0.163  &  0.160   \\

        Aug\_KL 
        & 0.080 & 0.073 & 0.060  &  0.043
         & 0.133&  0.070& 0.050  &  0.060 \\
        
    \text{MaxR\'enyi} 
    &  0.100& 0.210 & 0.163  &  0.127
         &  0.077  & 0.073  &  0.213 & 0.183 \\
        \midrule

    \rowcolor{gray!15} \textbf{Ours} (Image\_Sim, Gaussian Blur, Kernel Size 5) & \multicolumn{4}{c}{0.333}  & \multicolumn{4}{c}{0.323}  \\
    
    \rowcolor{gray!15} \textbf{Ours} (Image\_Sim, Motion Blur Kernel Size 5) & \multicolumn{4}{c}{0.363}  & \multicolumn{4}{c}{0.297}  \\

    \rowcolor{gray!15} \textbf{Ours} (Image\_Sim, JPEG Compression, Quality = 5) & \multicolumn{4}{c}{0.057}  & \multicolumn{4}{c}{0.060}  \\
    
        \bottomrule
    \end{tabular}
    }
    \caption{\textbf{TPR at 5\% FPR} of various baseline methods under \citet{LiWCTAC24}'s pipeline and our proposed approach on VL-MIA/Flickr. The column 'img', 'inst', 'desp', and 'inst+desp' has the same meaning as the previous table.}
    \label{tab:vl-mia-tpr}
\end{table*}

\noindent\textbf{Implementation Details.} All these baselines need the knowledge of the target model's tokenizer and the output logits. The implementations of all baselines are based on~\citet{LiWCTAC24}. Following the recommendation of~\citet{Min-K} and~\citet{LiWCTAC24}, we set $K = 20$ for the Min-K\% method. $K$ is set to 0, 10, and 100 for MaxR\'enyi-K\% and $\alpha$ is chosen over 0.5, 1, and 2 for both ModR\'enyi and MaxR\'enyi-K\%, as in~\citet{LiWCTAC24}. We report the highest AUC for each method and provide the TPR at 5\% FPR using the hyperparameter combination that achieves this highest AUC. For the Image Similarity-based attack, we fix the Kernel Size of the Gaussian blur and the Motion blur as 5. For JPEG compression, the image quality is set to 5 (lower values indicate stronger compression). For the Text Similarity-based attack, following~\citet{LiWCTAC24}, we use ``\textit{Describe this image concisely}''~\citep{LiWCTAC24} as the prompt and the max generation token amount is 32. For simplicity, we only use Gaussian blur for Text Similarity-based attack and the kernel size is set to 45.

\begin{figure}[htbp]
\centering 
\includegraphics[width=1\linewidth]{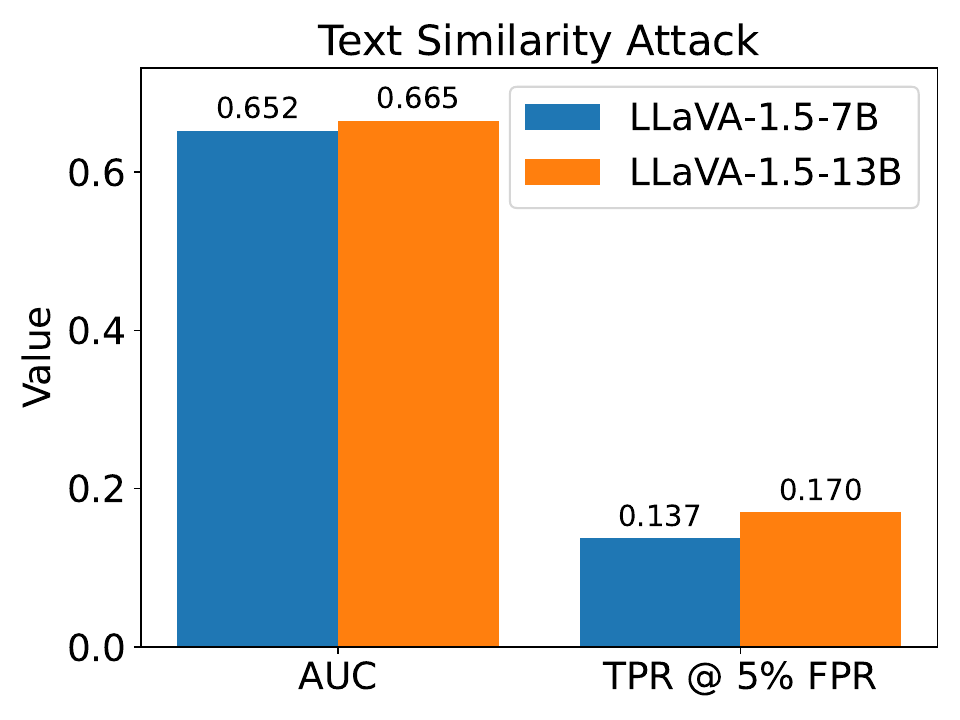}
\caption{Text similarity-based membership inference attack on VL-MIA/Flickr.} 
\label{fig:text_attack}
\end{figure}

\begin{figure*}[htbp]
\centering 
\includegraphics[width=1\linewidth]{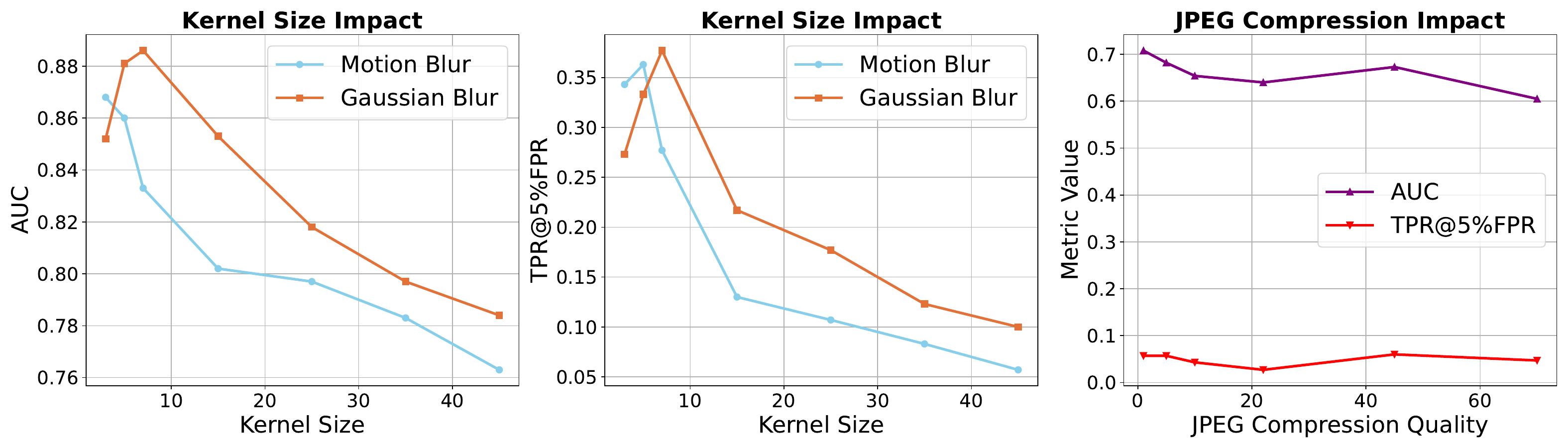}
\caption{Analysis on Image Corruption Hyperparameters.} 
\label{fig:kernel_size}
\end{figure*}

\subsection{White-Box MIA Performance}

The results on LLaVA 1.5 series are shown in Table~\ref{tab:vl-mia-auc} and Table~\ref{tab:vl-mia-tpr}.  Results on VL-MIA/Flickr 2K are in Appendix~\ref{app:2k}. \textbf{\textit{Please note that our methods have different knowledge of the models compared with the baseline methods in Table~\ref{tab:vl-mia-auc} and Table~\ref{tab:vl-mia-tpr}.}} These baseline methods can be viewed grey-box attack methods. We can obtain the image embeddings while they can obtain the output text's logits. We can observe that similarity-based attacks using Gaussian Blur and Motion Blur can achieve an AUC higher than 0.8 for both LLaVA 1.5-7B and LLaVA-1.5-13B, which is much higher than all the baselines. In comparison, the highest AUC among the baseline methods is 0.743 on LLaVA-1.5-7B and 0.738 on LLaVA-1.5-13B. Similar results can be found in Table~\ref{tab:vl-mia-tpr}, our methods largely outperform all the baselines in terms of TPR at 5\% FPR. Our best method can have a TPR at 5\% FPR higher than 0.3 on both models while the best baseline performance is 0.210 on LLaVA-1.5-7B and 0.227 on LLaVA-1.5-13B.

\begin{figure}[htbp]
\centering 
\includegraphics[width=1\linewidth]{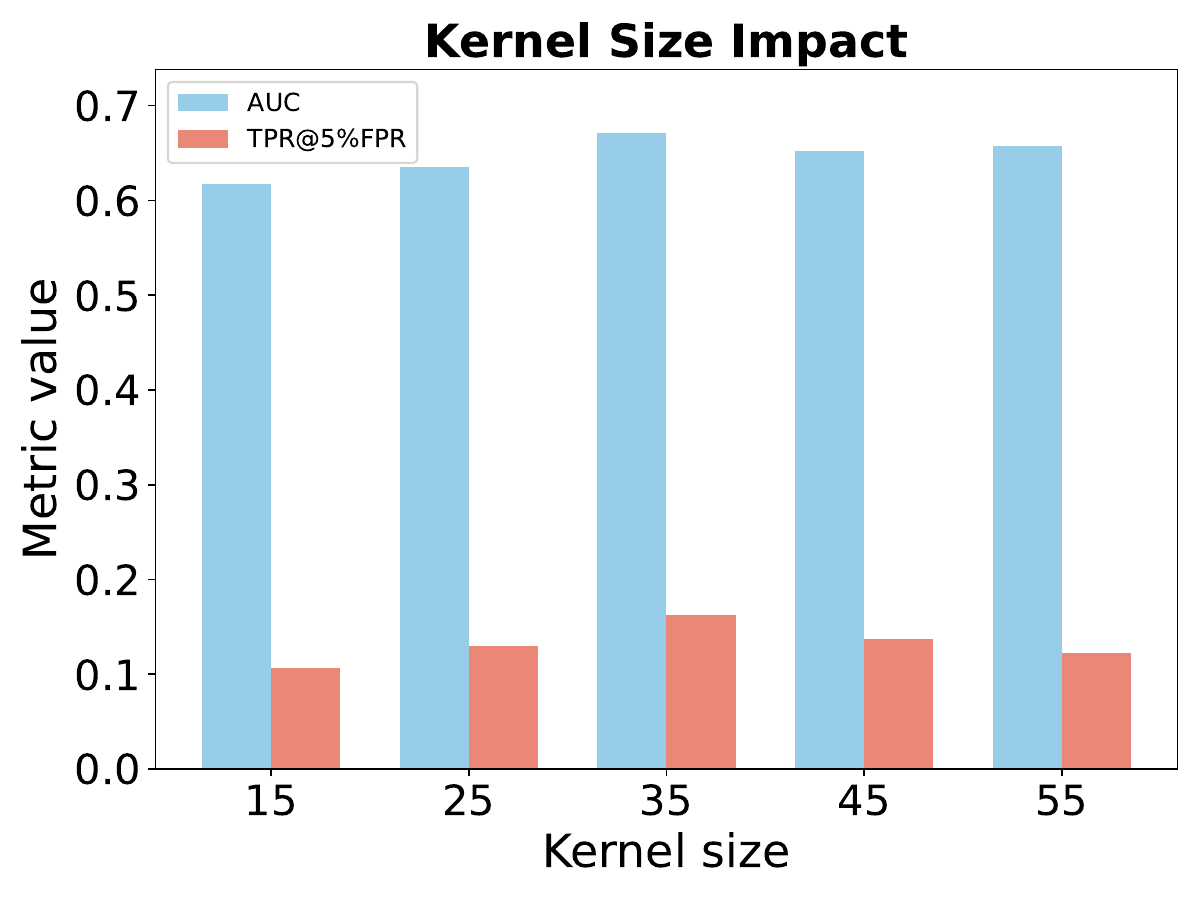}
\caption{Kernel Size's Impact on Text Similarity-based Attack.} 
\label{fig:kernel_size_text}
\end{figure}

The results show that using information from the visual side can better facilitate image MIAs compared with using information from different logit slices. The results also suggest that the choice of image corruption method plays an important role. For example, the performance using JPEG compression is worse than that using Gaussian Blur and Motion Blur.

\subsection{Black-box MIA Performance}
The results of our attack method under the black-box setting, which is based on output texts' embeddings, are shown in Figure~\ref{fig:text_attack}. The results demonstrate the effectiveness of our method. For example, it achieves an AUC of 0.652 and a TPR at 5\% FPR of 0.137 on LLaVA-1.5-7B, which outperforms many baselines in Table~\ref{tab:vl-mia-auc} that require using output logits, even though our approach only queries the target model and utilizes the output text.

\begin{table}[hbpt]
    \centering
    \resizebox{1\linewidth}{!}{
    \begin{tabular}{lcc|cc}
        \toprule
        \multirow{2}{*}{\textbf{Method}} 
        & \multicolumn{2}{c|}{\textbf{AUC}} 
        & \multicolumn{2}{c}{\textbf{TPR at 5\% FPR}} \\
        \cmidrule(lr){2-3} \cmidrule(lr){4-5}
        & 7B & 13B & 7B & 13B \\
        \midrule
        Ours (Gaussian Blur)         & 0.881 & 0.878 & 0.333 & 0.323 \\
        Ours (Gaussian Blur, CLIP)   & \multicolumn{2}{c|}{0.885} & \multicolumn{2}{c}{0.413} \\
        \hdashline
        Ours (Motion Blur)           & 0.860 & 0.856 & 0.363 & 0.297 \\
        Ours (Motion Blur, CLIP)     & \multicolumn{2}{c|}{0.858} & \multicolumn{2}{c}{0.393} \\

                \hdashline
        Ours (JPEG Compression)           & 0.682 & 0.681 & 0.057 & 0.060 \\
        Ours (JPEG Compression, CLIP)     & \multicolumn{2}{c|}{0.705} & \multicolumn{2}{c}{0.103} \\
        \bottomrule
    \end{tabular}
    }
    \caption{Image MIA \textbf{AUC} and \textbf{TPR at 5\% FPR} of our proposed approach on VL-MIA/Flickr. The embeddings are replaced with the ones obtained directly from the CLIP vision encoder without the alignment module.}
    \label{tab:vl-mia-auc-clip}
\end{table}

\subsection{Hyperparameter Analysis}
In this subsection, we conduct experiments to observe the impacts of hyperparameters.
For the image similarity-based attack, we vary the kernel size for Gaussian Blur and Motion Blur using values of 3, 5, 7, 15, 25, 35, and 45. The image quality for JPEG compression is selected across 1, 5, 10, 20, 45, and 70. The selected model is LLaVA 1.5 7B. The results are shown in Figure~\ref{fig:kernel_size}. We can observe that smaller kernel sizes generally contribute to better performance in terms of both AUC and TPR at 5\% FPR. Larger kernel sizes will bring stronger corruption. Therefore, for blur-based corruption methods, it suggests that a smaller corruption level could have better performance. Although the performance drops a lot with very large kernel sizes, it still outperforms most baselines that rely on output logits. For the third sub-figure, we can see that a smaller JPEG quality (Stronger compression) generally leads to higher AUC.

For the text similarity-based attack, the kernel size for Gaussian Blur is varied across 15, 25, 35, 45 and 55. The results are shown in Figure~\ref{fig:kernel_size_text}. Unlike image similarity-based attacks, text similarity-based attacks generally perform better with larger kernel sizes. One possible reason is that if the kernel size is too small, small changes in the image embeddings lead to even smaller changes in the textual output. This makes the difference between member and non-member images less clear. Since we are using the AUC as the metric, we don’t need to explicitly set the threshold $\lambda$.

\subsection{Extra Findings}

We replace the embeddings obtained by the CLIP vision encoder and the alignment module with the embeddings directly from the CLIP vision encoder without passing them through the alignment module. The results are shown in Table~\ref{tab:vl-mia-auc-clip}. Interestingly, we find that using the embeddings directly from the CLIP vision encoder can even have slightly better performance. However, the CLIP encoder is frozen during the training stage of LLaVA-1.5. This suggests that the images might be included in the CLIP model's original training data. Our method's strong performance might also benefit from this. This suggests that we need some new benchmark datasets to better define the image MIA problem in the context of LVLMs.

\section{Conclusion}
In this paper, we design novel membership inference attack methods named ICIMIA against LVLMs under both white-box and black-box settings. Our approach is based on the observation that LVLMs exhibit varying sensitivity to image corruption for member and non-member images. We leverage this phenomenon by using the pair-wise similarity of the original version and its corrupted counterpart as the metric. Experimental results on representative datasets validate the effectiveness of our proposed methods in image membership inference attacks.

\section*{Limitations}
We observed an interesting phenomenon: many non-member images generated by DALL·E 2~\citep{ramesh2022dalle2} exhibit greater robustness to corruption compared to their original counterparts (the generated image is prompted to be similar to its original member image), regardless of whether the original image is a member or non-member. Therefore, our method does not work for such a dataset where each non-member image is generated by DALL-E based on the original member image. One example is the VL-MIA/DALL-E dataset~\citep{LiWCTAC24} where Blip~\citep{li2023blip} generates a caption for each member image and the caption is used by DALL-E to generate a non-member image for this image. We leave this as our future work.

\section*{Ethical considerations}
Our work aims to find the images that are in the training data of large vision-language models. Our proposed method, ICIMIA, can help individuals know whether their sensitive data is used to train the model. All experiments are conducted on open-source models and publicly available datasets. In this paper, we only use AI assistants for grammar checking and sentence polishing.

\section*{Acknowledgments}
This material is based upon work supported by, or in part by the Army Research Office (ARO) under grant number W911NF-2110198, the Department of Homeland Security (DHS) under grant number 17STCIN00001-05-00, and Cisco Faculty Research Award. The findings in this paper do not necessarily reflect the view of the funding agencies.

\bibliography{custom}

@article{touvron2023llama,
  title={Llama: Open and efficient foundation language models},
  author={Touvron, Hugo and Lavril, Thibaut and Izacard, Gautier and Martinet, Xavier and Lachaux, Marie-Anne and Lacroix, Timoth{\'e}e and Rozi{\`e}re, Baptiste and Goyal, Naman and Hambro, Eric and Azhar, Faisal and Rodriguez, Aurelien and Joulin, Armand and Grave, Edouard and Lample, Guillaume },
  journal={arXiv preprint arXiv:2302.13971},
  year={2023}
}

@article{touvron2023llama2,
  title={Llama 2: Open foundation and fine-tuned chat models},
    author={Hugo Touvron and Louis Martin and Kevin Stone and Peter Albert and Amjad Almahairi and Yasmine Babaei and Nikolay Bashlykov and Soumya Batra and Prajjwal Bhargava and Shruti Bhosale and Dan Bikel and Lukas Blecher and Cristian Canton Ferrer and Moya Chen and Guillem Cucurull and David Esiobu and Jude Fernandes and Jeremy Fu and Wenyin Fu and Brian Fuller and Cynthia Gao and Vedanuj Goswami and Naman Goyal and Anthony Hartshorn and Saghar Hosseini and Rui Hou and Hakan Inan and Marcin Kardas and Viktor Kerkez and Madian Khabsa and Isabel Kloumann and Artem Korenev and Punit Singh Koura and Marie-Anne Lachaux and Thibaut Lavril and Jenya Lee and Diana Liskovich and Yinghai Lu and Yuning Mao and Xavier Martinet and Todor Mihaylov and Pushkar Mishra and Igor Molybog and Yixin Nie and Andrew Poulton and Jeremy Reizenstein and Rashi Rungta and Kalyan Saladi and Alan Schelten and Ruan Silva and Eric Michael Smith and Ranjan Subramanian and Xiaoqing Ellen Tan and Binh Tang and Ross Taylor and Adina Williams and Jian Xiang Kuan and Puxin Xu and Zheng Yan and Iliyan Zarov and Yuchen Zhang and Angela Fan and Melanie Kambadur and Sharan Narang and Aurelien Rodriguez and Robert Stojnic and Sergey Edunov and Thomas Scialom},
  journal={arXiv preprint arXiv:2307.09288},
  year={2023}
}

@article{tong2024cambrian,
  title={Cambrian-1: A fully open, vision-centric exploration of multimodal llms},
  author={Shengbang Tong and Ellis Brown and Penghao Wu and Sanghyun Woo and Manoj Middepogu and Sai Charitha Akula and Jihan Yang and Shusheng Yang and Adithya Iyer and Xichen Pan and Austin Wang and Rob Fergus and Yann LeCun and Saining Xie},
  journal={arXiv preprint arXiv:2406.16860},
  year={2024}
}

@inproceedings{liu2023llava,
 author = {Liu, Haotian and Li, Chunyuan and Wu, Qingyang and Lee, Yong Jae},
 booktitle = {Advances in Neural Information Processing Systems},
 pages = {34892--34916},
 title = {Visual Instruction Tuning},
 volume = {36},
 year = {2023}
}

@InProceedings{liu2024llava1.5,
    author    = {Liu, Haotian and Li, Chunyuan and Li, Yuheng and Lee, Yong Jae},
    title     = {Improved Baselines with Visual Instruction Tuning},
    booktitle = {Proceedings of the IEEE/CVF Conference on Computer Vision and Pattern Recognition (CVPR)},
    year      = {2024},
    pages     = {26296-26306}
}

@InProceedings{radford2021clip,
  title = 	 {Learning Transferable Visual Models From Natural Language Supervision},
  author =       {Radford, Alec and Kim, Jong Wook and Hallacy, Chris and Ramesh, Aditya and Goh, Gabriel and Agarwal, Sandhini and Sastry, Girish and Askell, Amanda and Mishkin, Pamela and Clark, Jack and Krueger, Gretchen and Sutskever, Ilya},
  booktitle = 	 {Proceedings of the 38th International Conference on Machine Learning},
  pages = 	 {8748--8763},
  year = 	 {2021},
}

@InProceedings{chen2024internvl,
    author    = {Chen, Zhe and Wu, Jiannan and Wang, Wenhai and Su, Weijie and Chen, Guo and Xing, Sen and Zhong, Muyan and Zhang, Qinglong and Zhu, Xizhou and Lu, Lewei and Li, Bin and Luo, Ping and Lu, Tong and Qiao, Yu and Dai, Jifeng},
    title     = {InternVL: Scaling up Vision Foundation Models and Aligning for Generic Visual-Linguistic Tasks},
    booktitle = {Proceedings of the IEEE/CVF Conference on Computer Vision and Pattern Recognition (CVPR)},
    year      = {2024},
    pages     = {24185-24198}
}

@misc{vicuna2023,
    title = {Vicuna: An Open-Source Chatbot Impressing GPT-4 with 90\%* ChatGPT Quality},
    url = {https://lmsys.org/blog/2023-03-30-vicuna/},
    author = {Chiang, Wei-Lin and Li, Zhuohan and Lin, Zi and Sheng, Ying and Wu, Zhanghao and Zhang, Hao and Zheng, Lianmin and Zhuang, Siyuan and Zhuang, Yonghao and Gonzalez, Joseph E. and Stoica, Ion and Xing, Eric P.},
    month = {March},
    year = {2023}
}

@article{bai2023qwenvl,
  title={Qwen-VL: A Versatile Vision-Language Model for Understanding, Localization, Text Reading, and Beyond},
  author={Bai, Jinze and Bai, Shuai and Yang, Shusheng and Wang, Shijie and Tan, Sinan and Wang, Peng and Lin, Junyang and Zhou, Chang and Zhou, Jingren},
  journal={arXiv preprint arXiv:2308.12966},
  year={2023}
}

@inproceedings{
zhu2024minigpt,
title={Mini{GPT}-4: Enhancing Vision-Language Understanding with Advanced Large Language Models},
author={Deyao Zhu and Jun Chen and Xiaoqian Shen and Xiang Li and Mohamed Elhoseiny},
booktitle={The Twelfth International Conference on Learning Representations},
year={2024}
}

@article{dong2024benchmarking,
  title={Benchmarking and Improving Detail Image Caption},
  author={Dong, Hongyuan and Li, Jiawen and Wu, Bohong and Wang, Jiacong and Zhang, Yuan and Guo, Haoyuan},
  journal={arXiv preprint arXiv:2405.19092},
  year={2024}
}

@inproceedings{changpinyo2021conceptual,
   author    = {Changpinyo, Soravit and Sharma, Piyush and Ding, Nan and Soricut, Radu},
    title     = {Conceptual 12M: Pushing Web-Scale Image-Text Pre-Training To Recognize Long-Tail Visual Concepts},
    booktitle = {Proceedings of the IEEE/CVF Conference on Computer Vision and Pattern Recognition (CVPR)},
    month     = {June},
    year      = {2021},
    pages     = {3558-3568}
}

@misc{kakaobrain2022coyo-700m,
  title         = {COYO-700M: Image-Text Pair Dataset},
  author        = {Byeon, Minwoo and Park, Beomhee and Kim, Haecheon and Lee, Sungjun and Baek, Woonhyuk and Kim, Saehoon},
  year          = {2022},
  howpublished  = {\url{https://github.com/kakaobrain/coyo-dataset}},
}

@article{hu2022miasurvey,
author = {Hu, Hongsheng and Salcic, Zoran and Sun, Lichao and Dobbie, Gillian and Yu, Philip S. and Zhang, Xuyun},
title = {Membership Inference Attacks on Machine Learning: A Survey},
year = {2022},
volume = {54},
journal = {ACM Computing Surveys},
}

@inproceedings{shokri2017miaml,
  title={Membership inference attacks against machine learning models},
  author={Shokri, Reza and Stronati, Marco and Song, Congzheng and Shmatikov, Vitaly},
  booktitle={2017 IEEE symposium on security and privacy (SP)},
  pages={3--18},
  year={2017}
}

@inproceedings{
oren2024proving,
title={Proving Test Set Contamination in Black-Box Language Models},
author={Yonatan Oren and Nicole Meister and Niladri S. Chatterji and Faisal Ladhak and Tatsunori Hashimoto},
booktitle={The Twelfth International Conference on Learning Representations},
year={2024}
}

@inproceedings{
duan2024membership,
title={Do Membership Inference Attacks Work on Large Language Models?},
author={Michael Duan and Anshuman Suri and Niloofar Mireshghallah and Sewon Min and Weijia Shi and Luke Zettlemoyer and Yulia Tsvetkov and Yejin Choi and David Evans and Hannaneh Hajishirzi},
booktitle={Conference on Language Modeling},
year={2024}
}

@inproceedings{LiWCTAC24,
 author = {Li, Zhan and Wu, Yongtao and Chen, Yihang and Tonin, Francesco and Abad Rocamora, Elias and Cevher, Volkan},
 booktitle = {Advances in Neural Information Processing Systems},
 pages = {98645--98674},
 title = {Membership Inference Attacks against Large Vision-Language Models},
 year = {2024}
}

@InProceedings{ko2023practical,
    author    = {Ko, Myeongseob and Jin, Ming and Wang, Chenguang and Jia, Ruoxi},
    title     = {Practical Membership Inference Attacks Against Large-Scale Multi-Modal Models: A Pilot Study},
    booktitle = {Proceedings of the IEEE/CVF International Conference on Computer Vision (ICCV)},
    year      = {2023},
    pages     = {4871-4881}
}

@article{fu2024mmevalsurvey,
  title={MME-Survey: A Comprehensive Survey on Evaluation of Multimodal LLMs},
  author={Chaoyou Fu and Yi-Fan Zhang and Shukang Yin and Bo Li and Xinyu Fang and Sirui Zhao and Haodong Duan and Xing Sun and Ziwei Liu and Liang Wang and Caifeng Shan and Ran He},
  journal={arXiv preprint arXiv:2411.15296},
  year={2024}
}

@misc{liu2024llavanext,
    title={LLaVA-NeXT: Improved reasoning, OCR, and world knowledge},
    url={https://llava-vl.github.io/blog/2024-01-30-llava-next/},
    author={Liu, Haotian and Li, Chunyuan and Li, Yuheng and Li, Bo and Zhang, Yuanhan and Shen, Sheng and Lee, Yong Jae},
    month={January},
    year={2024}
}

@inproceedings{nasr2019comprehensive,
  title={Comprehensive privacy analysis of deep learning: Passive and active white-box inference attacks against centralized and federated learning},
  author={Nasr, Milad and Shokri, Reza and Houmansadr, Amir},
  booktitle={2019 IEEE symposium on security and privacy (SP)},
  pages={739--753},
  year={2019}
}

@inproceedings{rezaei2021difficultymia,
    author    = {Rezaei, Shahbaz and Liu, Xin},
    title     = {On the Difficulty of Membership Inference Attacks},
    booktitle = {Proceedings of the IEEE/CVF Conference on Computer Vision and Pattern Recognition (CVPR)},
    year      = {2021},
    pages     = {7892-7900}
}

@inproceedings{leino2020stolen,
  title={Stolen Memories: Leveraging Model Memorization for Calibrated White-Box Membership Inference},
  author={Leino, Klas and Fredrikson, Matt},
  booktitle={29th USENIX security symposium (USENIX Security)},
  pages={1605--1622},
  year={2020}
}

@inproceedings{sablayrolles2019white,
  title={White-box vs black-box: Bayes optimal strategies for membership inference},
  author={Sablayrolles, Alexandre and Douze, Matthijs and Schmid, Cordelia and Ollivier, Yann and Jégou, Hervé },
  booktitle = 	 {Proceedings of the 36th International Conference on Machine Learning},
  pages = 	 {5558--5567},
  year={2019}
}

@article{salem2018ml,
  title={Ml-leaks: Model and data independent membership inference attacks and defenses on machine learning models},
  author={Salem, Ahmed and Zhang, Yang and Humbert, Mathias and Berrang, Pascal and Fritz, Mario and Backes, Michael},
  journal={arXiv preprint arXiv:1806.01246},
  year={2018}
}

@inproceedings{mattern2023mianeighbourhood,
    title = "Membership Inference Attacks against Language Models via Neighbourhood Comparison",
    author = {Mattern, Justus  and
      Mireshghallah, Fatemehsadat  and
      Jin, Zhijing  and
      Sch{\"o}lkopf, Bernhard  and
      Sachan, Mrinmaya  and
      Berg-Kirkpatrick, Taylor},
    booktitle = "Findings of the Association for Computational Linguistics: ACL 2023",
    year = "2023",
    pages = "11330--11343"
}

@inproceedings{mireshghallah2022empirical,
  title={An empirical analysis of memorization in fine-tuned autoregressive language models},
  author={Mireshghallah, Fatemehsadat and Uniyal, Archit and Wang, Tianhao and Evans, David K and Berg-Kirkpatrick, Taylor},
  booktitle={Proceedings of the 2022 Conference on Empirical Methods in Natural Language Processing},
  pages={1816--1826},
  year={2022}
}

@inproceedings{hu2022m,
 author = {Hu, Pingyi and Wang, Zihan and Sun, Ruoxi and Wang, Hu and Xue, Minhui},
 booktitle = {Advances in Neural Information Processing Systems},
 pages = {1867--1882},
title = {M${^4}$I: Multi-modal Models Membership Inference},

volume = {35},
 year = {2022}
}

@inproceedings{
Min-K,
title={Detecting Pretraining Data from Large Language Models},
author={Weijia Shi and Anirudh Ajith and Mengzhou Xia and Yangsibo Huang and Daogao Liu and Terra Blevins and Danqi Chen and Luke Zettlemoyer},
booktitle={The Twelfth International Conference on Learning Representations},
year={2024}
}

@article{lin2024decoding,
  title={Decoding Time Series with LLMs: A Multi-Agent Framework for Cross-Domain Annotation},
  author={Lin, Minhua and Chen, Zhengzhang and Liu, Yanchi and Zhao, Xujiang and Wu, Zongyu and Wang, Junxiang and Zhang, Xiang and Wang, Suhang and Chen, Haifeng},
  journal={arXiv preprint arXiv:2410.17462},
  year={2024}
}

@article{xu2024gnnllm,
  title={Llm and gnn are complementary: Distilling llm for multimodal graph learning},
  author={Xu, Junjie and Wu, Zongyu and Lin, Minhua and Zhang, Xiang and Wang, Suhang},
  journal={arXiv preprint arXiv:2406.01032},
  year={2024}
}

@article{wu2025lanp,
  title={LanP: Rethinking the Impact of Language Priors in Large Vision-Language Models},
  author={Wu, Zongyu and Niu, Yuwei and Gao, Hongcheng and Lin, Minhua and Zhang, Zhiwei and Zhang, Zhifang and Shi, Qi and Wang, Yilong and Fu, Sike and Xu, Junjie and Ao, Junjie and Dai, Enyan and Feng, Lei and Zhang, Xiang and Wang, Suhang},
  journal={arXiv preprint arXiv:2502.12359},
  year={2025}
}

@InProceedings{yue2023mmmu,
    author    = {Yue, Xiang and Ni, Yuansheng and Zhang, Kai and Zheng, Tianyu and Liu, Ruoqi and Zhang, Ge and Stevens, Samuel and Jiang, Dongfu and Ren, Weiming and Sun, Yuxuan and Wei, Cong and Yu, Botao and Yuan, Ruibin and Sun, Renliang and Yin, Ming and Zheng, Boyuan and Yang, Zhenzhu and Liu, Yibo and Huang, Wenhao and Sun, Huan and Su, Yu and Chen, Wenhu},
    title     = {MMMU: A Massive Multi-discipline Multimodal Understanding and Reasoning Benchmark for Expert AGI},
    booktitle = {Proceedings of the IEEE/CVF Conference on Computer Vision and Pattern Recognition (CVPR)},
    year      = {2024},
    pages     = {9556-9567}
}

@article{li2024seedbenchplus,
  title={Seed-bench-2-plus: Benchmarking multimodal large language models with text-rich visual comprehension},
  author={Li, Bohao and Ge, Yuying and Chen, Yi and Ge, Yixiao and Zhang, Ruimao and Shan, Ying},
  journal={arXiv preprint arXiv:2404.16790},
  year={2024}
}

@inproceedings{carlini2019secret,
  title={The secret sharer: Evaluating and testing unintended memorization in neural networks},
  author={Carlini, Nicholas and Liu, Chang and Erlingsson, {\'U}lfar and Kos, Jernej and Song, Dawn},
  booktitle={Proceedings of the 28th USENIX Conference on Security Symposium},
  pages={267--284},
  year={2019}
}

@article{hu2025membership,
  title={Membership Inference Attacks Against Vision-Language Models},
  author={Hu, Yuke and Li, Zheng and Liu, Zhihao and Zhang, Yang and Qin, Zhan and Ren, Kui and Chen, Chun},
  journal={arXiv preprint arXiv:2501.18624},
  year={2025}
}

@inproceedings{liu2021encodermi,
  title={EncoderMI: Membership Inference against Pre-trained Encoders in Contrastive Learning},
  author={Liu, Hongbin and Jia, Jinyuan and Qu, Wenjie and Gong, Neil Zhenqiang},
  booktitle={Proceedings of the 2021 ACM SIGSAC Conference on Computer and Communications Security},
  pages={2081--2095},
  year={2021}
}

@article{wang2024comprehensive,
author = {Wang, Fali and Zhang, Zhiwei and Zhang, Xianren and Wu, Zongyu and Mo, TzuHao and Lu, Qiuhao and Wang, Wanjing and Li, Rui and Xu, Junjie and Tang, Xianfeng and He, Qi and Ma, Yao and Huang, Ming and Wang, Suhang},
title = {A Comprehensive Survey of Small Language Models in the Era of Large Language Models: Techniques, Enhancements, Applications, Collaboration with LLMs, and Trustworthiness},
year = {2025},
volume = {16},
journal = {ACM Transactions on Intelligent Systems and Technology}
}

@article{zhang2023llamaadapter,
  title={Llama-adapter: Efficient fine-tuning of language models with zero-init attention},
  author={Zhang, Renrui and Han, Jiaming and Liu, Chris and Gao, Peng and Zhou, Aojun and Hu, Xiangfei and Yan, Shilin and Lu, Pan and Li, Hongsheng and Qiao, Yu},
  journal={arXiv preprint arXiv:2303.16199},
  year={2023}
}

@inproceedings{ren2024self,
author = {Ren, Jie and Chen, Kangrui and Chen, Chen and Sehwag, Vikash and Xing, Yue and Tang, Jiliang and Lyu, Lingjuan},
title = {Self-Comparison for Dataset-Level Membership Inference in Large (Vision-)Language Model},
year = {2025},
booktitle = {Proceedings of the ACM on Web Conference 2025},
pages = {910–920},
}

@InProceedings{li2023blip,
  title = {{BLIP}-2: Bootstrapping Language-Image Pre-training with Frozen Image Encoders and Large Language Models},
  author = {Li, Junnan and Li, Dongxu and Savarese, Silvio and Hoi, Steven},
  booktitle = {Proceedings of the 40th International Conference on Machine Learning},
  pages = 	 {19730--19742},
  year = 	 {2023}
}

@inproceedings {carlini2021extracting,
author = {Nicholas Carlini and Florian Tram{\`e}r and Eric Wallace and Matthew Jagielski and Ariel Herbert-Voss and Katherine Lee and Adam Roberts and Tom Brown and Dawn Song and {\'U}lfar Erlingsson and Alina Oprea and Colin Raffel},
title = {Extracting Training Data from Large Language Models},
booktitle = {30th USENIX Security Symposium (USENIX Security)},
year = {2021},
pages = {2633--2650}
}

@inproceedings{li2021miadinclassify,
  title={Membership inference attacks and defenses in classification models},
  author={Li, Jiacheng and Li, Ninghui and Ribeiro, Bruno},
  booktitle={Proceedings of the Eleventh ACM Conference on Data and Application Security and Privacy},
  pages={5--16},
  year={2021}
}

@inproceedings{
zhang2025mink++,
title={Min-K\%++: Improved Baseline for Pre-Training Data Detection from Large Language Models},
author={Jingyang Zhang and Jingwei Sun and Eric Yeats and Yang Ouyang and Martin Kuo and Jianyi Zhang and Hao Frank Yang and Hai Li},
booktitle={The Thirteenth International Conference on Learning Representations},
year={2025}
}

@inproceedings{renyi1961measures,
  title={On measures of entropy and information},
  author={R{\'e}nyi, Alfr{\'e}d},
  booktitle={Proceedings of the fourth Berkeley symposium on mathematical statistics and probability, volume 1: contributions to the theory of statistics},
  year={1961},
  organization={University of California Press}
}

@article{ramesh2022dalle2,
  title={Hierarchical text-conditional image generation with clip latents},
  author={Ramesh, Aditya and Dhariwal, Prafulla and Nichol, Alex and Chu, Casey and Chen, Mark},
  journal={arXiv preprint arXiv:2204.06125},
  year={2022}
}

@inproceedings{lin2014mscoco,
  title={Microsoft coco: Common objects in context},
  author="Lin, Tsung-Yi
    and Maire, Michael
    and Belongie, Serge
    and Hays, James
    and Perona, Pietro
    and Ramanan, Deva
    and Doll{\'a}r, Piotr
    and Zitnick, C. Lawrence",
    editor="Fleet, David
    and Pajdla, Tomas
    and Schiele, Bernt
    and Tuytelaars, Tinne",
  booktitle={Computer Vision -- ECCV 2014},
  pages={740--755},
  year={2014}
}

@article{opencv_library,
    author = {Bradski, G.},
    journal = {Dr. Dobb's Journal of Software Tools},
    title = {{The OpenCV Library}},
    year = {2000}
}

@article{bucciarelli2024personalizing,
  title={Personalizing multimodal large language models for image captioning: an Experimental analysis},
  author={Bucciarelli, Davide and Moratelli, Nicholas and Cornia, Marcella and Baraldi, Lorenzo and Cucchiara, Rita},
  journal={arXiv preprint arXiv:2412.03665},
  year={2024}
}

@inproceedings{song2017machine,
  title={Machine Learning Models that Remember Too Much},
  author={Song, Congzheng and Ristenpart, Thomas and Shmatikov, Vitaly},
  booktitle={Proceedings of the 2017 ACM SIGSAC Conference on Computer and Communications Security},
  pages={587--601},
  year={2017}
}

@inproceedings{
carlini2023quantifying,
title={Quantifying Memorization Across Neural Language Models},
author={Nicholas Carlini and Daphne Ippolito and Matthew Jagielski and Katherine Lee and Florian Tramer and Chiyuan Zhang},
booktitle={The Eleventh International Conference on Learning Representations},
year={2023}
}

@InProceedings{choo2021labelmia,
  title = 	 {Label-Only Membership Inference Attacks},
  author =       {Choquette-Choo, Christopher A. and Tramer, Florian and Carlini, Nicholas and Papernot, Nicolas},
  booktitle = 	 {Proceedings of the 38th International Conference on Machine Learning},
  pages = 	 {1964--1974},
  year = 	 {2021}
}

@inproceedings{li2023medllava,
 author = {Li, Chunyuan and Wong, Cliff and Zhang, Sheng and Usuyama, Naoto and Liu, Haotian and Yang, Jianwei and Naumann, Tristan and Poon, Hoifung and Gao, Jianfeng},
 booktitle = {Advances in Neural Information Processing Systems},
 pages = {28541--28564},
 title = {LLaVA-Med: Training a Large Language-and-Vision Assistant for Biomedicine in One Day},
 year = {2023}
}

@inproceedings{carlini2022membership,
  title={Membership inference attacks from first principles},
  author={Carlini, Nicholas and Chien, Steve and Nasr, Milad and Song, Shuang and Terzis, Andreas and Tramer, Florian},
  booktitle={2022 IEEE symposium on security and privacy (SP)},
  pages={1897--1914},
  year={2022}
}

@article{zhao2023survey,
    title={A Survey of Large Language Models},
    author={Zhao, Wayne Xin and Zhou, Kun and Li, Junyi and Tang, Tianyi and Wang, Xiaolei and Hou, Yupeng and Min, Yingqian and Zhang, Beichen and Zhang, Junjie and Dong, Zican and Du, Yifan and Yang, Chen and Chen, Yushuo and Chen, Zhipeng and Jiang, Jinhao and Ren, Ruiyang and Li, Yifan and Tang, Xinyu and Liu, Zikang and Liu, Peiyu and Nie, Jian-Yun and Wen, Ji-Rong},
    year={2023},
    journal={arXiv preprint arXiv:2303.18223},
}

@article{chen2023minigptv2,
  title={Minigpt-v2: large language model as a unified interface for vision-language multi-task learning},
  author={Chen, Jun and Zhu, Deyao and Shen, Xiaoqian and Li, Xiang and Liu, Zechun and Zhang, Pengchuan and Krishnamoorthi, Raghuraman and Chandra, Vikas and Xiong, Yunyang and Elhoseiny, Mohamed},
  journal={arXiv preprint arXiv:2310.09478},
  year={2023}
}

@InProceedings{li2024monkey,
    author    = {Li, Zhang and Yang, Biao and Liu, Qiang and Ma, Zhiyin and Zhang, Shuo and Yang, Jingxu and Sun, Yabo and Liu, Yuliang and Bai, Xiang},
    title     = {Monkey: Image Resolution and Text Label Are Important Things for Large Multi-modal Models},
    booktitle = {Proceedings of the IEEE/CVF Conference on Computer Vision and Pattern Recognition (CVPR)},
    year      = {2024},
    pages     = {26763-26773}
}

@misc{2023internlm,
    title={InternLM: A Multilingual Language Model with Progressively Enhanced Capabilities},
    author={InternLM Team},
    howpublished = {\url{https://github.com/InternLM/InternLM-techreport}},
    year={2023}
}

@article{penedo2023refinedweb,
      title={The RefinedWeb Dataset for Falcon LLM: Outperforming Curated Corpora with Web Data, and Web Data Only}, 
      author={Guilherme Penedo and Quentin Malartic and Daniel Hesslow and Ruxandra Cojocaru and Alessandro Cappelli and Hamza Alobeidli and Baptiste Pannier and Ebtesam Almazrouei and Julien Launay},
      year={2023},
        journal={arXiv preprint arXiv:2306.01116},
}

@article{chen2024internvl15,
  title={How far are we to gpt-4v? closing the gap to commercial multimodal models with open-source suites},
  author={Zhe Chen and Weiyun Wang and Hao Tian and Shenglong Ye and Zhangwei Gao and Erfei Cui and Wenwen Tong and Kongzhi Hu and Jiapeng Luo and Zheng Ma and Ji Ma and Jiaqi Wang and Xiaoyi Dong and Hang Yan and Hewei Guo and Conghui He and Botian Shi and Zhenjiang Jin and Chao Xu and Bin Wang and Xingjian Wei and Wei Li and Wenjian Zhang and Bo Zhang and Pinlong Cai and Licheng Wen and Xiangchao Yan and Min Dou and Lewei Lu and Xizhou Zhu and Tong Lu and Dahua Lin and Yu Qiao and Jifeng Dai and Wenhai Wang},
  journal={Science China Information Sciences},
  volume={67},
  year={2024},
  publisher={Springer}
}

@inproceedings{zhang2025divideandrefine,
    title = "Divide-Verify-Refine: Can {LLM}s Self-align with Complex Instructions?",
    author = "Zhang, Xianren  and
      Tang, Xianfeng  and
      Liu, Hui  and
      Wu, Zongyu  and
      He, Qi  and
      Lee, Dongwon  and
      Wang, Suhang",
    booktitle = "Findings of the Association for Computational Linguistics: ACL 2025",
    year = "2025",
    pages = "13783--13800",
}

\appendix
\label{sec:appendix}

\begin{table*}[htbp]
  \centering
  \begin{tabular}{lcc}
    \toprule
    Name & VL-MIA/Flickr & VL-MIA/Flickr 2K \\
    \midrule
    Source of Member Data  & MS COCO & MS COCO \\
    Source of Non-member Data & Flickr  & Flickr  \\
    \#Member Data          & 300     & 1000    \\
    \#Non-member Data      & 300     & 1000    \\
    \bottomrule
  \end{tabular}
  \caption{Dataset statistics. The datasets are constructed by~\citet{LiWCTAC24}.}
  \label{tab:stat}
\end{table*}

\begin{table*}
    \centering

    \resizebox{1\linewidth}{!}{
    \begin{tabular}{lcccc}
        \toprule
        \multirow{2}{*}{\textbf{Method}} 
        & \multicolumn{2}{c}{\textbf{AUC}} 
        & \multicolumn{2}{c}{\textbf{TPR at 5\% FPR}} \\
        \cmidrule(lr){2-3} \cmidrule(lr){4-5}
        & \textbf{LLaVA-1.5-7B} & \textbf{LLaVA-1.5-13B} 
        & \textbf{LLaVA-1.5-7B} & \textbf{LLaVA-1.5-13B} \\
        \midrule
        \rowcolor{gray!15} \textbf{Ours} (Image\_Sim, Gaussian Blur, Kernel Size 5) 
        & 0.854 & 0.850 & 0.349 & 0.337 \\                   
        \rowcolor{gray!15} \textbf{Ours} (Image\_Sim, Motion Blur, Kernel Size 5) 
        & 0.842 & 0.837 & 0.372 & 0.352 \\   
        \rowcolor{gray!15} \textbf{Ours} (Image\_Sim, JPEG Compression, Quality = 5) 
        & 0.629 & 0.635 & 0.088 & 0.080 \\    
        \bottomrule
    \end{tabular}
    }
    \caption{Image MIA performance of our proposed approach on VL-MIA/Flickr 2K.}
    \label{tab:vl-mia-2k}
\end{table*}

\newpage
\section{Details of Datasets}
\label{app:datasets}
Many models, such as LLaVA 1.5~\citep{liu2024llava1.5} and Minigpt 4~\citep{zhu2024minigpt}, use MS COCO~\citep{lin2014mscoco} to train the models. Therefore,~\citet{LiWCTAC24} use part of the images in MS COCO as the member images. For non-member data, \citet{LiWCTAC24} select images uploaded to Flickr\footnote{https://www.flickr.com/} after those target models' release date. VL-MIA/Flickr and VL-MIA/Flickr 2K are licensed under the Creative Commons Attribution 4.0 International Public License. The dataset statistics are summarized in Table~\ref{tab:stat}.

\section{Used Metrics}
\label{app:metric}
We provide details about two used metrics here:

\begin{itemize}
    \item \textbf{AUC}: Area Under the Curve (AUC) is the value of the area beneath the ROC curve. It is a widely used metric to evaluate the classification model's performance under all possible classification thresholds~\citep{LiWCTAC24}.
    
    \item \noindent \textbf{TPR at 5\% FPR}: True Positive Rate at a fiex False Positive Rate is another widely used metric to evaluate the performance of MIA methods~\citep{LiWCTAC24,carlini2022membership}. Following~\citet{LiWCTAC24}, we use TPR at 5\% FPR which reflects the value of the True Positive Rate when the False Positive Rate is 5\%.

\end{itemize}

\section{Additional Results on More Datasets}
\label{app:2k}
We conduct experiments on VL-MIA/Flickr-2K. The image similarity-based MIA results are shown in Table~\ref{tab:vl-mia-2k}. We can have similar observations as those shown in Table~\ref{tab:vl-mia-auc} and Table~\ref{tab:vl-mia-tpr}. The experimental results on VL-MIA/Flickr-2K also validate the effectiveness of our proposed methods.

\section{Potential Risks}
Although our proposed ICIMIA is designed to protect data safety and privacy, malicious users can use ICIMIA to infer whether an image is used to train a certain LVLM and then get some private information. For example, if the attacker knows that one person's medical image is used to train an LVLM for a certain disease, the attacker can get the information that this person might have this disease.

\end{document}